\documentclass[11pt]{article}
\usepackage[T1]{fontenc}
\usepackage{graphicx}
\usepackage{hyperref}
\usepackage{color}

\usepackage{amssymb, amsmath}
\usepackage[table]{xcolor}
\usepackage{graphicx} % Required for inserting images
\usepackage{mathtools}
\usepackage{nicefrac}
\usepackage{bbold}

\newcommand\cella{\cellcolor{yellow!20} a_{1,1}}
\newcommand\cellb{\cellcolor{green!50} a_{1,2}}
\newcommand\cellc{\cellcolor{red!40} a_{2,1}}
\newcommand\celld{\cellcolor{blue!60} a_{2,2}}

\newtheorem{claim}{Claim}[section]
\newtheorem{theorem}{Theorem}[section]

\begin{document}
%
% \title{A Polynomial Interpretation of Simplified Residual Neural Networks}
\title{Algebraic Representations for Faster Predictions
in Convolutional Neural Networks\thanks{Supported by the 
National Science Foundation under grant DMS 1854513.}}
%
% \titlerunning{Algebraic Representations for Faster Predictions in CNNs}
\author{Johnny Joyce\thanks{University of Illinois at Chicago,
Department of Mathematics, Statistics, and Computer Science,
851 S. Morgan St. (m/c 249), Chicago, IL 60607-7045, U.S.A.
Email: {\tt jjoyce22@uic.edu}, URL: {\tt https://jjoyce22.people.uic.edu/}.}
 \and
Jan Verschelde\thanks{University of Illinois at Chicago,
Department of Mathematics, Statistics, and Computer Science,
851 S. Morgan St. (m/c 249), Chicago, IL 60607-7045, U.S.A.
Email: {\tt janv@uic.edu}, URL: {\tt http://www.math.uic.edu/$\sim$jan}.}}
%
%\authorrunning{J. Joyce and J. Verschelde}
%\institute{University of Illinois at Chicago \\
%Department of Mathematics, Statistics, and Computer Science \\
%851 S Morgan St (m/c 249), Chicago, IL, 60607 \\
%\email{\{jjoyce22,janv\}@uic.edu}}
%
\maketitle              % typeset the header of the contribution
\begin{abstract}
Convolutional neural networks (CNNs) are a popular choice of model
for tasks in computer vision.  When CNNs are made with many layers,
resulting in a \emph{deep} neural network, skip connections may be 
added to create an easier gradient optimization problem 
while retaining model expressiveness. 
In this paper, we show that arbitrarily complex, trained, linear CNNs
with skip connections can be simplified into a single-layer model,
resulting in greatly reduced computational requirements during prediction time.
We also present a method for training nonlinear models with skip connections
that are gradually removed throughout training, giving the benefits of skip
connections without requiring computational overhead during during 
prediction time. These results are demonstrated with practical examples 
on Residual Networks (ResNet) architecture.

\end{abstract}


\section{Introduction}

% \textbf{Goal: Explain skip connections with polynomial models.}

Convolutional neural networks (CNNs) are a type of multi-layer neural 
network that primarily utilize convolution (cross-correlation) operations,
as opposed to regular neural networks (NNs),
which utilize fully-connected layers. 
Like fully-connected layers, convolutions can be represented as 
transformation matrices \cite{kohn2022geometry}, 
but possess different properties that make them well-suited for tasks 
involving images and videos, such as classification and segmentation.

CNNs, along with other types of neural networks, have the ability to represent functions whose complexity grows exponentially as the number of layers in the network increases \cite{raghu2017expressive}. \emph{Deep} neural networks, which consist of many layers, therefore offer great performance enhancements over their shallow counterparts, but are more difficult to train if considerations are not made on how to account for this tradeoff. \emph{Skip connections} (or shortcut connections or residual connections) help make NNs easier to train --- these work by taking the output of a particular layer, then adding it to the input of some later layer.
That is, if a neural network takes input $X$, and the $i^\text{th}$ layer computes $f(X)$, and the $(j-1)^{th}$ layer computes $g(X)$, then a skip connection from layer $i$ to layer $j$ gives $f(X)+g(X)$ as the input to the $j^\text{th}$ layer.
A famous example of a model that uses skip connections is \texttt{ResNet} \cite{he2016deep}, whose common variants may have up to 152 layers, and have won various machine learning competitions.

Generally speaking, skip connections allow for the training process to be an easier optimization problem by retaining information as it moves throughout the network and by allowing gradients to backpropagate through the network more easily, mitigating the `vanishing gradients' problem. More specific explanations are available, such as that they remove singularities in the loss landscape \cite{orhan2018skip}, but a full algebraic explanation has not yet been formulated.

\subsection{machine learning and algebraic geometry}

% \textbf{TODO: Look for things related more to computer algebra than algebraic geometry. Put 2.1 into intro.}

Algebraic geometry has been used in various ways to analyze neural networks (NNs) and machine learning in general. In 2018, Zhang et al \cite{zhang2018tropical} made connections between NNs and tropical geometry by showing that NNs with ReLU activation functions can be represented as tropical signomial maps. Using this connection, they gave bounds on number of regions of a NN's decision boundary, showed that zonotopes form ``building blocks'' of NNs under a tropical geometry view, and showed that the expressiveness of NNs grows exponentially with the number of layers. This sparked interest in this recent field was surveyed by Maragos et al in 2022 \cite{maragos2021tropical}. 

Algebraic interpretations of neural networks have also taken different approaches. In 2022, Kohn et al \cite{kohn2022geometry} examined the function space of linear CNNs, concluding that training a linear CNN with gradient descent results in a bias towards repeated filters that are not globally optimal. This, along with other examples \cite{laurent2018deep,kileel2019expressive}, is one example of a deep field of work where the loss landscape of NNs is examined --- a topic that is also often the focus of \textit{singular learning theory}. One such work by Hardt and Ma in 2017 \cite{hardt2017identity} directly uses such an analysis of loss landscapes to show that deep residual neural networks, which use skip connections, have no critical points other than global minima.

Singular learning theory is another approach that provides rich connections between algebraic geometry, as explored in great detail in Watanabe's 2009 book \cite{watanabe2009algebraic}. Examples of important results from this field include that almost all learning machines (NNs, mixture models, etc) are singular, so regular statistical learning theory does not apply \cite{watanabe2007almost} (thereby motivating why analyzing NNs is important); and that optimizing Bayes cross-validation is equivalent to optimizing another criterion used in singular learning theory \cite{watanabe2010asymptotic}. Such work allowed for exact computation of marignal likelihood integrals (model evidence) for certain models \cite{lin2011algebraic} where otherwise it would have usually been approximated.

CNNs, skip connections (as in residual NNs), and singularities were tied together by Orhan and Pitkow in 2018 \cite{orhan2018skip}, who showed that the use of skip connections in a CNN avoids singularities previously analyzed by Wei et al \cite{wei2008dynamics}. However, it should be noted that this approach does not \textit{resolve} existing singularities (e.g. as one would by applying techniques like successive blow-ups), but instead demonstrates that architectures with skip connections are not subject to singularities in the first place, thereby differing from approaches like those of Watanabe in singular learning theory. Conversely, the other direction has been explored, with NNs having been used to; resolve singularites \cite{berczi2023ml}, analyze algebraic structures \cite{bao2023neurons}, solve problems in ways inspired by algebraic geometric methods \cite{huang2022hompinns}, and solve problems while maintaining awareness of singularities \cite{hu2023solving}. Symbolic computation also has rich connections with machine learning, including the use of polynomials to analyze neural networks \cite{kohn2022geometry}, and applied machine learning to assist 
in symbolic computation 
\cite{florescu2024constrained,pickering2024explainable}.

\subsection{contributions and structure}

The structure of the remainder of this paper is as follows:

\begin{itemize}
    \item In Section~\ref{sec:cnnsetup}, we introduce a setup and relevant terminology for \emph{linear} CNNs (LCNs), with emphasis on viewing convolutions as transformation matrices.

    \item In Section~\ref{sec:skipconnections}, we add the necessary framework for skip connections to our setup and present a theorem that allows for the prediction function arbitrarily complex trained LCNs with skip connections to be pre-calculated. This facilitates the creation of models with strong performance on image-related tasks that only require the resources of a single-layer perceptron when making predictions. This can give an arbitrarily high speedup factor --- in the case of a linearized version of \texttt{ResNet34}, we observe a speedup of 98\%.

    \item In Section~\ref{sec:removeskipconnections}, we expand our view to include nonlinear CNNs, and present a method for removing skip connections, resulting in an observed 22\% to 46\% speedup in prediction time in practical experiments.

    \item In Section~\ref{sec:removeskipconnections}, we conclude by exploring potential future research directions based on our results.
\end{itemize}

All code used in this paper's experiments is available in a Jupyter Notebook file stored a public GitHub repository \footnote{\url{https://github.com/johnnyvjoyce/simplify-skip-connections}}.
% All code used in this paper's experiments is available in a public GitHub repository \footnote{\url{https://github.com/johnnyvjoyce/simplify-skip-connections}}.

\section{CNN setup}\label{sec:cnnsetup}

In this section, we describe a setup for \emph{feed-forward} CNNs --- we use feed-forward to distinguish networks that do not contain skip connections. This setup is similar to that of Kohn et al. \cite{kohn2022geometry}, who describe 1-dimensional convolutions as Toeplitz matrices. 
% In this section, the differences from the setup of Kohn et al.\ are that we focus on 2-dimensional inputs with 1-dimensional representations, and that we focus on classification rather than regression. Unlike Kohn et al., we also assume that convolutions come with an associated bias. We will expand upon this setup in Section~\ref{sec:skipconnections} to allow for skip connections, facilitating the use of much larger and deeper networks.
Our setup for this section differs from that of Kohn et al.\ in the following ways;
we focus on 2-dimensional (or higher) inputs and represent them as 1-dimensional vectors, 
we focus primarily on classification rather than regression (though this setup can also be applied to regression), and
we assume that all convolution operations are immediately followed by adding a corresponding bias vector.
Furthermore, we will expand upon this setup in Section~\ref{sec:skipconnections} to allow for skip connections, facilitating the use of much larger and deeper neural networks.

We start with a basic CNN architecture consisting of $L-1$ convolutional layers, followed by a single fully-connected layer. 
For each convolutional layer, indexed by $i=1,\dots,L-1$, we use $\alpha^{(i)}$ to refer to the layer as a function from its input space to its output space, $K^{(i)}$ to refer to its kernel, its stride as $s^{(i)}$, and its bias as $B^{(i)}$. The fully-connected layer $\alpha^{(L)}$ allows us to perform classification by mapping the output of the final convolutional layer to a vector of size $C$, where $C$ corresponds to the number of class labels. For each $i\in\{1,\dots,C\}$ the $i^\text{th}$ entry of the output corresponds to the probability that the input is associated with the $i\text{th}$ class. 

It is common to use a \emph{softmax} activation function after the final layer (which maps the output vector to a probability distribution), but to maintain linearity, we instead treat the softmax layer as part of the loss function. This does not change the optimization problem, and therefore has no positive or negative effect on model performance.

Inputs to our CNN are two-dimensional grayscale images, which would ordinarily be best represented by a ($h\times w$)-matrix $M$ containing saturation values ranging from 0 (black) to 1 (white). However, our notation and calculations involving convolutions are made simpler if we first ``unravel" our input into a vector $X$ of size $hw$, where $X_i = M_{(\lfloor \nicefrac{i}{w} \rfloor , \, i \text{ mod } w)}$ for all $i\in\{1,\dots,hw\}$.
We may also use color images in RGB format, which would contain 3 saturation values for each pixel --- one for each of red, green, and blue. This scenario is analogous, but requires unraveling a tensor of rank 3 rather than a matrix. We can take arbitrarily high input dimensions for other scenarios by unravelling in this way if needed.

With this setup, we can view all of the CNN's layers $\{\alpha^{(i)}:i=1,\dots,n\}$ as affine maps $\alpha^{(i)}(x)=W^{(i)} X + B^{(i)}$ --- the convolutional layers are viewed as Toeplitz matrices, while the fully-connected layer keeps its usual representation. The CNN can then be represented as a map $f(X) = (\alpha^{(n)} \circ \dots \circ \alpha^{(1)}) (X)$. Furthermore, we have the following:

\begin{eqnarray}
    f(X) & = & (\alpha^{(L)} \circ \dots \circ \alpha^{(1)}) (X)\\
    & = & W^{(L)} ( \dots ( W^{(2)} (W^{(1)} X + B^{(1)}) + B^{(2)} ) \dots ) + B^{(L)} \\
    & = & W^{(L)} ( \dots ( W^{(2)} W^{(1)} X + W^{(2)} B^{(1)} + B^{(2)} ) \dots ) + B^{(L)}\\
    & = & \Big(W^{(L)} \dots W^{(2)} W^{(1)}\Big) X \nonumber \\ 
    &  & + \Big( (W^{(L)}\dots W^{(2)})B^{(1)} + \dots + W^{(2)} B^{(n-1)} + B^{(L)} \Big)\\
    & = & \Big( \prod_{i=1}^{L} W^{(i)} \Big) X \;\; + \;\; \sum_{i=1}^{L} \Big( ( \prod_{j=i+1}^{L} W^{(j)} ) B^{(i)} \Big)
\end{eqnarray}\label{eq:cnnaffine}

Hence $f(X)$ is itself an affine map, given by $f(X)=W X+B$, where $\displaystyle{W\coloneqq\prod_{i=1}^{L} W^{(i)}}$ and $\displaystyle{B\coloneqq \sum_{i=1}^{L} \big( ( \prod_{j=i+1}^{L} W^{(j)} ) B^{(i)} \big)}$ (where $\prod$ uses left-multiplication of successive matrices).

We may also consider the map $g \coloneqq \alpha^{(n-1)} \circ \dots \circ \alpha^{(1)}$, which consists of all convolutional layers, with the final fully-connected layer removed. This is a map from our inputs into \emph{latent space}, which is a lower-dimensional representation of the original inputs. We have that $g$ is also an affine map, given by $g(X)=W_\text{latent} X + B_\text{latent}$, where $\displaystyle{W_\text{latent}\coloneqq\prod_{i=1}^{n-1} W^{(i)}}$ and $\displaystyle{B_\text{latent}\coloneqq \sum_{i=1}^{L-1} \big( ( \prod_{j=i+1}^{L-1} W^{(j)} ) B^{(i)} \big)}$.

\section{Skip Connections in Linear Networks}\label{sec:skipconnections}

Skip connections add the outputs of multiple layers together as information moves through a neural network. To achieve this, the operands for these tensor addition operations must be the same size. However, convolutional layers in a CNN \emph{reduce} the size of an image if no counteractive measures are taken.

Therefore, we introduce \emph{resampling} and \emph{padding} in Section~\ref{subsec:resampling}, both of which allow for the loss in size when using convolutions to be counteracted, making them a necessary prerequisite for adding skip connections. These are both standard operations in machine learning, but we show them to emphasize a somewhat uncommon approach of representing them as transformation matrices. In Section~\ref{subsec:skip}, we utilize these transformation matrices to create a network with skip connections.

Though padding and resampling both achieve similar goals, they provide different and complementary utility. Padding an image before performing a convolution ensures that the output of the convolution has the same size as the input before padding --- a CNN that only utilizes padding will therefore perform successive convolutions on an image that retains its size. On the other hand, resampling allows us to explicitly choose the size of the input, making it either smaller or larger --- a CNN that only utilizes resampling will have inputs that get smaller as successive convolutions are applied. By manipulating the size of the image as it passes through layers, we can alter the \emph{receptive field} of the network.

% \textbf{TODO: Explain that transformation matrices are needed for theorem 1.}

% \textbf{TODO: Explain that padding is good for keeping same size, whereas resampling allows either local or global features to be detected.}

% \textbf{TODO: Cite Barzilai, receptive field, condition numbers --- more of a numerical algebra approach}

% In this section, we iterate upon our framework by introducing skip connections.
% To achieve this, we first need to introduce matrices that can be used to re-sample and to pad images, which are introduced in Section~\ref{subsec:resampling}. In Section~\ref{subsec:skip}, we utilize these matrices to create a network with skip connections.

% These operations are necessary for skip connections, which add the output of different layers. Vector addition requires vectors to have the same dimensions, but each convolutional layer reduces the image 

\subsection{Transformation matrices for resampling and padding}\label{subsec:resampling}

Take a $(2\times2)-$matrix $A\coloneqq\left[\begin{array}{cc}
\cella & \cellb \\
\cellc & \celld 
\end{array}\right]$. Each of the entries have been assigned an arbitrary color for ease of visualizing this matrix as an image.

\subsubsection{Resampling.}\label{subsubsec:resampling}

Suppose we want to resample $A$ into a $(3\times4)-$matrix using nearest-neighbor interpolation. To achieve this, we first need to reshape $A$ into a $(4\times 1)$ column vector. We can then use the transformation matrix shown on the left-hand side of (\ref{eq:resampling}) to obtain the $(12\times 1)$ column vector shown on the right-hand side of (\ref{eq:resampling}). 

\begin{align}\label{eq:resampling}
\begin{bmatrix}
1 & 0 & 0 & 0\\ 
1 & 0 & 0 & 0\\ 
0 & 1 & 0 & 0\\ 
0 & 1 & 0 & 0\\ 
0 & 0 & 1 & 0\\ 
0 & 0 & 1 & 0\\ 
0 & 0 & 0 & 1\\ 
0 & 0 & 0 & 1\\ 
0 & 0 & 1 & 0\\ 
0 & 0 & 1 & 0\\ 
0 & 0 & 0 & 1\\ 
0 & 0 & 0 & 1
\end{bmatrix}
\cdot
\left[
\begin{array}{c}
\cella \\
\cellb \\
\cellc \\
\celld 
\end{array}
\right]
=
\left[
\begin{array}{c}
\cella \\
\cella \\
\cellb \\
\cellb \\
\cellc \\
\cellc \\
\celld \\
\celld \\ 
\cellc \\
\cellc \\
\celld \\
\celld 
\end{array}
\right]
\end{align}

Finally, by reshaping the resulting column vector into a $(3\times4)-$matrix, we obtain:

\begin{align}
B=\left[
\begin{array}{cccc}
\cella & \cella & \cellb & \cellb \\
\cellc & \cellc & \celld & \celld \\
\cellc & \cellc & \celld & \celld
\end{array}
\right]
\end{align}

By comparing the colors of the original matrix $A$ against the colors of the result $B$, we can see that this resampling has preserved the positions of matrix entries as they would appear if they were pixels in an image.

In this example, the second row of $B$ contains entries from the final row of $A$, but an equally valid approach could have been to take these entries from the first row of $A$. Alternatively, if we wished to instead use a more complex resampling method, such as bilinear interpolation, we could achieve this by changing the entries of our transformation matrix in (\ref{eq:resampling}). 
% For bilinear sampling, each row of the transformation matrix should still sum to 1, but the nonzero entries would correspond to the positions of each of the nearest neighbors of each entry of the output.

For simplicity, we use nearest-neighbor interpolation in our experiments, and for consistency, entries of our resampling matrix are found through coordinate transforms in a way that matches the implementation of the common Python packages Pillow, scikit-image, and PyTorch.

\subsubsection{Padding.}\label{subsubsec:padding}

Another common technique in CNNs is \emph{padding}, where a border is introduced around the edges of an image. These borders may consist of zeros, samples of pixels near the border, or other methods. 

For example, if we use the sample example matrix $A$ as in Section~\ref{subsubsec:resampling} and reshape it into a vector in the same way, we can pad $A$ with a border of zeros of size 1 with the following transformation matrix:

\begin{align}\label{eq:padding}
\begin{bmatrix}
0 & 0 & 0 & 0\\ 
0 & 0 & 0 & 0\\ 
0 & 0 & 0 & 0\\
0 & 0 & 0 & 0\\
0 & 0 & 0 & 0\\
1 & 0 & 0 & 0\\
0 & 1 & 0 & 0\\
0 & 0 & 0 & 0\\
0 & 0 & 0 & 0\\
0 & 0 & 1 & 0\\
0 & 0 & 0 & 1\\
0 & 0 & 0 & 0\\
0 & 0 & 0 & 0\\
0 & 0 & 0 & 0\\
0 & 0 & 0 & 0\\
0 & 0 & 0 & 0\\
\end{bmatrix}
\cdot
\left[
\begin{array}{c}
\cella \\
\cellb \\
\cellc \\
\celld 
\end{array}
\right]
=
\left[
\begin{array}{c}
0 \\ 0      \\ 0      \\ 0 \\
0 \\ \cella \\ \cellb \\ 0 \\
0 \\ \cellc \\ \celld \\ 0 \\
0 \\ 0      \\ 0      \\ 0
\end{array}
\right]
\end{align}

Like last time, we only need to reshape back into given dimensions 
to obtain the result:

\begin{align}
    C=\left[
    \begin{array}{cccc}
    0 & 0      & 0      & 0 \\
    0 & \cella & \cellb & 0 \\
    0 & \cellc & \celld & 0 \\
    0 & 0      & 0      & 0
    \end{array}
    \right]
\end{align}

If we were to now perform any convolution with kernel size $[2,2]$ on $C$, our output will have size $2\times 2$, which matches the original image. Hence, performing appropriately-sized padding before a convolution will preserve the dimensions of the input and output. 

In general, if we wish to perform a convolution with kernel size $(k_1,\dots k_n)$ on a tensor of order $n$ and we wish for the output shape to match the input shape, we need to pad each dimension $i$ with $k_i-1$ entries on each side.

\subsection{Skip connections}\label{subsec:skip}

With resampling matrices and padding matrices now introduced, we can create CNNs with arbitrarily many skip connections, which can have many more layers, allowing for greater expressiveness.

In Section~\ref{sec:cnnsetup}, we created feed-forward networks by composing matrices that represent successive layers and demonstrated that the result was an affine map. Here, we use a slightly different setup; instead of creating maps $\alpha^{(i)}$ that act as a single layer which we then compose together, we construct our network $f$ recursively by defining maps $f^{(1)}, \dots , f^{(n)}$. We then take $f=f^{(L)}$. These maps are defined as follows:

% \begin{align*}
%     f^{(0)}(X)=X
% \end{align*}

% \begin{align}\label{eq:skiprecursive}
%     f^{(i)}(X)=W^{(i)} P^{(i)} \left(\sum_{k=0}^{i-1} t^{(k,i-1)} R^{(k,i-1)} f^{(k)}(X) \right) + B^{(i)} \text{\; \; for all $i=1,\dots,L$}
% \end{align}

\begin{alignat}{4}\label{eq:skiprecursive}
    f^{(0)}(X) & = X \nonumber \\
    f^{(i)}(X) & = W^{(i)} P^{(i)} \left(\sum_{k=0}^{i-1} t^{(k,i-1)}  R^{(k,i-1)} f^{(k)}(X) \right) \mathrlap{  + B^{(i)} } \\
    & & \text{\; \; for all $i\in\{1,\dots,L\}$} \nonumber
\end{alignat}

Here, for all $i$, we take $\{t^{(k,i)} : k=0,\dots,i\}$ to be constants s.t. $\sum_{k=0}^{i} t^{(k,i)}=1$. These represent the weights of our skip connections, and we use $i-1$ as the second index since skip connections augment the \textit{inputs} to each layer. For any pairs of layers $k$ and $i$ where we do not desire a skip connection, we simply set $t^{(k,i-1)}=0$. 

Note that we can also set $t^{(k,k)}=0$, allowing for cases where a layer does not feed into the subsequent layer. This facilitates neural networks with many `paths' (e.g. one path consisting of even-numbered layers and another for odd-numbered layers).

Each $R^{(k,i-1)}$ is a resampling matrix that resamples the dimensions of $f^{(k)}(X)$ to match the dimensions of $f^{(i-1)}(X)$ (which are the \textit{input} dimensions for $f^{(i)}(X)$). Each $P^{(i)}$ is a padding matrix.

In practice, not every layer of a CNN would need to use both resampling matrices and padding matrices, since using one negates the necessity for the other. For any $P^{(i)}$ or $R^{(k,i-1)}$ that we do not wish to use, we can replace them with identity matrices. To retain generality, we will keep all resampling matrices and padding matrices in our calculations.

We can also add \emph{batch normalization} into our toolkit, which scales inputs into each layer by multiplying and adding scalar values. This change is trivial, so we will ignore it in our equations for simplicity.

We can now make the following observation:

\begin{claim}\label{claim:cnnaffine}
    Let $f$ be a map corresponding to a linear CNN with arbitrarily many skip connections. Then $f(X)$ is an affine map.
\end{claim}

% \begin{proof}
{\em Proof.}
Induction on layers.

First, note that $f^{(0)}$ is clearly affine.

Next, assume that each of $f^{(0})$ through $f^{(i-1)}$ are affine for some $i\in \{1,\dots,L\}$. Then $f^{(i-1)}$ is given by (\ref{eq:skiprecursive}), and is an affine sum of affine maps, and is therefore affine.

\hfill Q.E.D.
% \end{proof}

This observation gives way to the following:

% \begin{theorem}\label{thm:simplifiedlinearcnn}
%     $\displaystyle{f(X) = f_{W}^{(L)}(\mathbb{1}_{hw}) X + f^{(L)}(\Vec{0})}$

%     ...where $\displaystyle{f^{(i)}(X) \coloneqq W^{(i)} P^{(i)} \left(\sum_{k=0}^{i-1} t^{(k,i-1)} R^{(k,i-1)} f^{(k)}(X) \right)}$ is the same as in (\ref{eq:skiprecursive}) with all bias terms removed.

% \end{theorem}

\begin{theorem}\label{thm:simplifiedlinearcnn}
    Let $\displaystyle{f_W^{(i)}(X) \coloneqq W^{(i)} P^{(i)} \left(\sum_{k=0}^{i-1} t^{(k,i-1)} R^{(k,i-1)} f^{(k)}(X) \right)}$ be the same as in (\ref{eq:skiprecursive}) with all bias terms removed. Also take $\mathbb{1}_{hw}$ to be the identity matrix of size $hw \times hw$, and take $\Vec{0}$ to be the zero vector of length $hw$. Then the following holds:
    
    \begin{equation*}
        \displaystyle{f(X) = f_{W}^{(L)}(\mathbb{1}_{hw}) X + f^{(L)}(\Vec{0})},
    \end{equation*}

    \noindent where $f$ is the map given by a CNN with $L$ layers and arbitrarily many skip connections, and where $X$ is an $h\times w$ input matrix that has been reshaped into a vector of length $hw$.
\end{theorem}

%\begin{proof}
{\em Proof.}

    Induction on layers $i=1,\dots,L$ to show that $f^{(i)}(X) = f_{W}^{(i)}(\mathbb{1}_{hw}) X + f^{(i)}(\Vec{0})$.
    
    First, take layer $i=1$. We have:

    \begin{align}
        f^{(1)}(\Vec{0}) &= W^{(1)} P^{(1)} \left( t^{(0,0)} R^{(0,0)} f^{(0)}(\Vec{0}) \right) + B^{(1)}  \\
        &= W^{(1)} P^{(1)} \left( t^{(0,0)} R^{(0,0)} \Vec{0} \right) + B^{(1)}  \\
        & = B^{(1)}
    \end{align}
    
    Hence:
    
    \begin{align}
        f^{(1)}(X) &= W^{(1)} P^{(1)} \left( t^{(0,0)} R^{(0,0)} f^{(0)}(X) \right) + B^{(1)} \\
        % &= W^{(2)} P^{(2)} \left( t^{(0,1)} R^{(0,1)} \mathbb{1}_{hw} \right) X  + B^{(2)} \\
        &= W^{(1)} P^{(1)} \left( t^{(0,0)} R^{(0,0)} f^{(0)}(\mathbb{1}_{hw})\right) X  + B^{(1)} \\
        &= f_W^{(1)}(\mathbb{1}_{hw}) X  + f^{(1)}(\Vec{0})
    \end{align}

    Next, assume that $f^{(k)}(X) = f_{W}^{(k)}(\mathbb{1}_{hw}) X + f^{(k)}(\Vec{0})$ for all $k\in\{1,\dots,i\}$. Then:

    \begin{align}
        f^{(i+1)}(X) &= W^{(i+1)} P^{(i+1)} \left(\sum_{k=1}^{i} t^{(k,i)} R^{(k,i)} f^{(k)}(X) \right) + B^{(i+1)} \\
        &= W^{(i+1)} P^{(i+1)} \left(\sum_{k=1}^{i} t^{(k,i)} R^{(k,i)} \left(f_{W}^{(k)}(\mathbb{1}_{hw}) X + f^{(k)}(\Vec{0})\right) \right) + B^{(i+1)} \\ 
        &= W^{(i+1)} P^{(i+1)} \left(\sum_{k=1}^{i} t^{(k,i)} R^{(k,i)} f_{W}^{(k)}(\mathbb{1}_{hw}) X \right)\\  
        & \;\;\;\;\; + W^{(i+1)} P^{(i+1)} \left(\sum_{k=1}^{i} t^{(k,i)} R^{(k,i)} f^{(k)}(\Vec{0}) \right) + B^{(i+1)} \\ 
        &= W^{(i+1)} P^{(i+1)} \left(\sum_{k=1}^{i} t^{(k,i)} R^{(k,i)} f_{W}^{(k)}(\mathbb{1}_{hw}) \right) X +  f^{(i+1)}(\Vec{0}) \\ 
        &= f_W^{(i+1)}(\mathbb{1}_{hw}) X +  f^{(i+1)}(\Vec{0})
    \end{align}

    % \begin{align*}
    %     f^{(L)}(X) &= W^{(L)} P^{(L)} \left(\sum_{k=1}^{L-1} t^{(k,L-1)} R^{(k,L-1)} f^{(k)}(X) \right) + B^{(L)} \\
    %     &= W^{(L)} P^{(L)} \left(\sum_{k=1}^{L-1} t^{(k,L-1)} R^{(k,L-1)} \left(f_W^{(k)}(X) - f^{(k)}(X)\right) \right) + B^{(L)} \\
    % \end{align*}

    Taking $i=L$, we obtain the theorem. \hfill Q.E.D.

%\end{proof}

Theorem~\ref{thm:simplifiedlinearcnn} allows us to pre-calculate $f_{W}^{(L)}(\mathbb{1}_{hw})$ and $f^{(L)}(\Vec{0})$, allowing us to make predictions far faster. When making predictions using a trained model, we need only calculate $f(X)=WX+B$, where $W=f_{W}^{(L)}(\mathbb{1}_{hw})$ and $B=f^{(L)}(\Vec{0})$. 

In practice, this allows us to use the predictive power of arbitrarily complex LCNs with skip connections while requiring the resources of a single-layer perceptron when the trained model makes predictions.

\section{Removing Skip Connections With a Homotopy}\label{sec:removeskipconnections}

%\textbf{TODO: Explain that this may circumvent singularities - either locally or in the loss function. Refer to Orhan 2018. Specify that this would be a future direction of research.}

In this section, we present another method of reducing the computational resources for making predictions with CNNs. Unlike in Section~\ref{sec:skipconnections}, this method applies to both linear and nonlinear networks.

% % When implementing skip connections, we perform an assignment operation $x \gets x^{(i)} + x$, where $x$ .
% A pseudocode implementation of a single skip connection may look like  $x \gets x^{(i)} + x$, where $x$ is the 

% $f^{(i)}(X) = \text{conv}\left(f^{(i-1)}(X) + f^{(j)}(X) \right) $
In (\ref{eq:skiprecursive}), we gave a recursive definition for skip connections. In standard CNNs with skip connections, one would set $t^{(i-1,i-1)}=0.5$ for all $i$, set $t^{(j,i-1)}=0.5$ for at most one value of $j<i-1$, and set $t^{(k,i-1)}=0$ otherwise. This would mean that the input to the $i^\text{th}$ layer would be given by $0.5 x^{(i-1)} + 0.5 x^{(j)}$, where $x^{(i-1)}$ and $x^{(j)}$ are the padded and scaled outputs of the $(i-1)^\text{th}$ and $j^\text{th}$ layers, respectively. 

We can simplify this scenario by unifying all of our $t^{(k,i-1)}$ values into a single parameter that we simply call $t$. 
We then set the input to the $i^\text{th}$ layer to: 

\begin{equation}
    t \cdot x^{(i-1)} + (1-t) \cdot x^{(j)}
\end{equation}

Note that when $t=0.5$, we get the standard setup for skip connections, and when $t=0$, we get a network with no skip connections. %and when $t=1$, we get a network with full weight on the skip connections and no weight on the previous layer. 
Hence, if we allow $t$ to be a parameter of our model $f$, we see that $f(x,t)$ is a homotopy between skip connection models and feed-forward models. This allows us to train a model that utilizes the benefit of skip connections by starting at some $t_0\in(0,1)$, then reduce $t$ as the model trains so that the final model is $f(x,0)$.

By having no skip connections in the resulting model, 
we obtain reduced computation times and memory usage when making predictions.

For homotopies to work out numerically,
the solution paths need to be free of singularities.
In the setup of neural networks, randomness is introduced
via the application of the stochastic gradient methods and we conjecture
that this removes the difficulties caused by singularities.
But this remains a subject of future study.

\section{Computational Experiments}\label{sec:experiments}

In Section~\ref{sec:skipconnections} and Section~\ref{sec:removeskipconnections}, we introduced methods for obtaining improved prediction times. In this section, we use practical experiments to evaluate the speedup obtained with these methods, which are split across Subsections~\ref{subsec:experimentsprecompute} and ~\ref{subsec:experimentsremoveskip}.

\subsection{Equipment, Software, and Data}

In both experiments, we use the MNIST database of handwritten digits with 60,000 labeled images for the training set, and 10,000 labeled images for the validation set. All models are built in PyTorch, and run on an NVIDIA GeForce RTX 2060 Mobile GPU. When we refer to prediction time, we record the mean wall clock time across an entire epoch of the dataset, without including any overhead (such as transforming data and loading the data into the GPU). We use a batch size of 64.

We also utilize \texttt{ResNet} -- in particular, \texttt{ResNet34}, named as such because it has 34 convolutional layers -- a model that is known to produce strong performance on image classification problems and can be trained relatively easily while having a deep architecture. \texttt{ResNet34} consists of `basic blocks,' which are two $3\times3$ convolutions. Multiple basic blocks form a `layer,' and there are four layers in total, plus one convolution before and after these layers. Each layer contains 3, 4, 6, and 3 basic blocks, respectively, giving a total of 34 convolutions, and each layer's output has a skip connection to the subsequent layer's input.

\subsection{Pre-computing CNNs}\label{subsec:experimentsprecompute}

\begin{figure}[htb]
    \centering
%    \includesvg[width=0.9\textwidth]{images/linear_performance_comparison.svg}
\centerline{\includegraphics[width=0.9\textwidth]{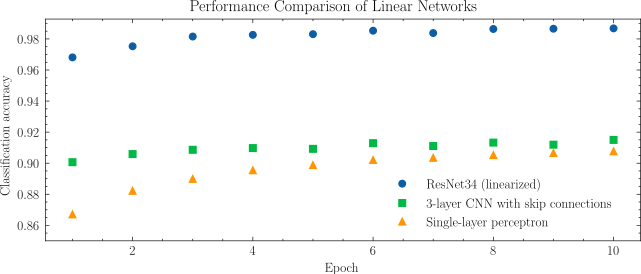}}
    \caption{Classification accuracy for varying values of $t$ (skip connection strength) in \text{ResNet34}. 10 epochs were performed over the MNIST dataset in each trial, and the mean result over 5 trials was taken for each point shown in the scatter plot.}
    \label{fig:linear_performance_comparison}
\end{figure}

In Section~\ref{sec:removeskipconnections}, we introduced a method to pre-compute CNNs with skip connections, reducing the computational requirements to that of a single-layer perceptron. To demonstrate the effectiveness of this reduction, Figure~\ref{fig:linear_performance_comparison} compares the accuracy of various linear networks on the validation set of MNIST as the networks train. 
One of these networks is a modified version of \texttt{ResNet34} with all nonlinear ReLU activation functions removed (which results in a linear network), and the other network is as described in Section~\ref{sec:skipconnections} with 3 convolutional layers and a skip connection from the first layer to the third layer.
We see that both the 3-layer network and the linearized \texttt{ResNet34} produce better performance than the single-layer perceptron without sacrificing speed or memory when making predictions.
%produce vastly different performance due to the more complex models being able to more effectively navigate the loss surface.

We can also measure the speedup compared to a version of each of these models that have not been pre-computed.
When averaged over 5 epochs, the mean prediction time for the linearized \texttt{ResNet34} for a single image is 104.91 microseconds.
For a basic 3-layer CNN with skip connections, the mean prediction time is 12.05 $\mu$s.
Since both of these models fit the description of Theorem~\ref{thm:simplifiedlinearcnn}, they can both be reduced to single-layer perceptrons, which have a mean prediction time of 2.06 $\mu$s. In the case of \texttt{ResNet34}, this is a speedup of over \textbf{98\%}.

\subsection{Practical experiment for removing skip connections}\label{subsec:experimentsremoveskip}

\begin{figure}[htb]
    \centering
%    \includesvg[width=0.9\textwidth]{images/t_vs_accuracy_resnet34.svg}
\centerline{\includegraphics[width=0.9\textwidth]{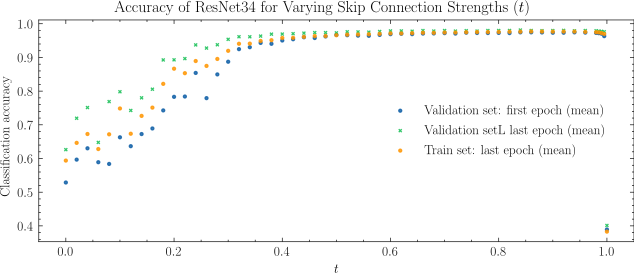}}
    \caption{Classification accuracy for varying values of $t$ (skip connection strength) in \text{ResNet34}. 2 epochs were performed over the MNIST dataset in each trial, and the mean result over 5 trials was taken for each point shown in the scatter plot.}
    \label{fig:t_vs_accuracy_resnet34}
\end{figure}

In this section, we compare the accuracy of a standard \texttt{ResNet34} against one that gradually has its skip connections removed, as described in Section~\ref{sec:removeskipconnections}. To achieve this, we need to set an initial value for $t$ (let us call the initial value $t_0$), which then decreases as the model trains, before reaching 0 at the end of training. To decide on a suitable $t_0$, we could either opt for the standard choice of 0.5, or we could refer to Figure~\ref{fig:t_vs_accuracy_resnet34}, which shows average classification accuracy on the validation set across the first two epochs for various choices of $t_0$. Using Figure~\ref{fig:t_vs_accuracy_resnet34}, we see that values near (but not equal to) 1 produce high initial classification accuracies on the validation set, so we use an initial value of $t_0=0.9$.

It is worth noting that the shape of the graph in Figure~\ref{fig:t_vs_accuracy_resnet34} differs for other networks. For shallow networks, we observe roughly constant performance for values of $t$ up to a certain threshold, before sharply dropping off --- this is because shallow networks do not benefit as much from skip connections as deeper networks, and greatly suffer when the strength of skip connections is too strong, resulting in weak performance for high $t$ values. On the other hand, for moderately deep networks, such as \texttt{ResNet18}, we observe an upside-down u-shaped curve, with moderate-strength skip connections performing best. Examples of such graphs can be seen in this project's Github repository. In our case, with \texttt{ResNet34}, we observe the best performance with very high values of $t$ because these allow the gradients to propagate through the network more easily, which is particularly important for deeper networks. $t=1$ produces the worst relative performance because it severely reduces the expressiveness of the model by cutting out the skipped layers altogether.

With our initial value chosen, we have many choices for how to schedule $t$ as we train. One simple option would be to set thresholds for validation accuracy that, when passed, result in a pre-specified, lower value of $t$. More advanced methods for scheduling other hyperparameters in deep learning, such as learning rate, have already been explored \cite{li2020exponential} and could be applied to $t$ too. However, to maintain simplicity and problem tractability, we opt for a method where we schedule different predetermined values of $t$ for each epoch, allowing us to have a fixed number of epochs. In our case, we use 10 epochs, with corresponding values of $t$ for each epoch set to \texttt{\{0.9, 0.7, 0.5, 0.3, 0.2, 0.1, 0.05, 0.025, 0.01, 0\}} accordingly.

\begin{figure}[hbt]
    \centering
%    \includesvg[width=0.9\textwidth]{images/static_vs_scheduled_t.svg}
\centerline{\includegraphics[width=0.9\textwidth]{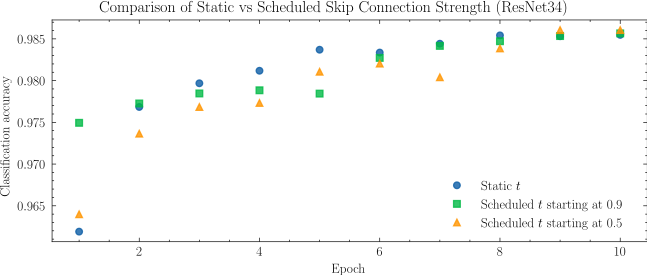}}
    \caption{Classification accuracy of \text{ResNet34} on MNIST validation set across 10 epochs using a ResNet34 model with \textit{(a)} fixed $t$ (skip connection strength) compared to \textit{(b)} scheduled values of $t$ that decrease to 0 \textit{(c)} scheduled values starting at 0.5 that can be applied to other models. Each point shown is the mean over 5 trials.}
    \label{fig:static_vs_scheduled_t}
\end{figure}

Figure~\ref{fig:static_vs_scheduled_t} shows a comparison across 10 epochs with scheduled $t$ values against a setup where we keep $t$ static at 0.5 (which gives the standard setup with skip connections), with results averaged over 5 trials. We see that the model trained with a schedule for $t$ does not significantly lose validation accuracy over the static model.

Figure~\ref{fig:static_vs_scheduled_t} also shows another schedule starting at $t_0=0.5$: \texttt{\{0.5, 0.4, 0.3, 0.2, 0.1, 0.05, 0.025, 0.01, 0.05, 0\}}. This schedule is more representative of a situation where we do not know what value of $t_0$ would work best --- this is often likely to be the case because it can be computationally expensive to produce results as in Figure~\ref{fig:t_vs_accuracy_resnet34}. This schedule can be applied searching for a good $t_0$, making it widely applicable. Again, we still do not observe any significant loss in accuracy.

To contextualize the speed difference when making predictions, we need only compare a version of \texttt{ResNet34} that uses skip connections against one that does not. When averaged over 5 epochs of the MNIST dataset placed in batches of 64 running on an NVIDIA GeForce RTX 2060 Mobile, the mean time for a single prediction with \texttt{ResNet34} is 106.83 microseconds. On the other hand, when skip connections are removed (by explicitly taking their programming out of the network rather than simply setting $t=0$), the average time is 83.1 microseconds. This is a speedup of approximately \textbf{22\%}.

The speedup is even more stark for networks with more skip connections. By creating a basic 6-layer CNN as described in Section~\ref{sec:cnnsetup}, a version with skip connections between all non-consecutive layers averages 30.72 $\mu$s, compared to an average of 16.44 $\mu$s with skip connections removed, a \textbf{46\%} speedup.

\section{Future Directions}

% While our focus in this paper has been on linear convolutional networks, LCNs do not fully leverage the predictive power of more general neural networks. To do so, we would need to include nonlinear activation functions in our setup, such as the ReLU function, which takes the identity function for positive values and sets negative values to zero. The use of activation functions results in non-linear maps, which are more expressive and become more complex with more layers \textbf{TODO: Cite exponential complexity increase in number of layers}

We have obtained two similar but distinct results; one showing that arbitrarily complex \emph{linear} CNNs can be vastly simplified, and another showing that nonlinear CNNs can be lightly simplified. Unifying these two results would provide great computational improvements while leveraging the full expressiveness of nonlinear CNNs.

To achieve this, one would need to represent nonlinear functions of matrices efficiently. ReLU, along with all other piecewise linear activation functions (e.g. leaky ReLU and step functions), have been explored by Zhang, Naitzat, and Lim \cite{zhang2018tropical}, who characterized neural networks using such activation functions with tropical geometry. By using this framework on LCNs with skip connections, one may be able to train nonlinear neural networks with skip connections that are later removed (as we have shown), before simplifying the network by pre-computing the transformation map from input space to output space. This could involve pre-computing the model's function on each of the distinct parts of function space into which the model divides the input plane into locally linear functions. In such a case, nonlinearities may need to be used sparingly to avoid dividing the plane an intractable number of times.

Another area of focus may be to apply more advanced methods of scheduling the skip connection strength parameter $t$. Previously-explored methods for scheduling the learning rate, parameter $\alpha$ (such as in \cite{li2020exponential}) could be adapted for use here with slight modifications, such as requiring that $t$ eventually becomes sufficiently close to 0. 
% Given that changing $t$ also changes the model itself, techniques from transfer learning may be relevant, such as techniques for
Alternatively, since changing $t$ also changes the model itself, we could view this approach as a form of \emph{knowledge distillation} \cite{gou2021knowledge}, one could use further techniques from this field to cut down the size of the network much more aggressively. This approach may even be combined with the tropical geometry approach to reduce the number of nonlinearities, resulting in a pre-computable and tractable model that preserves the teachers performance while reducing memory costs and computational costs.

%\begin{credits}
% \subsubsection{\ackname} A bold run-in heading in small font size at the end of the paper is
% used for general acknowledgments, for example: This study was funded
%\subsubsection{\discintname}
%The authors have no competing interests to declare that are relevant to the content of this article.
%%\end{credits}

\bibliographystyle{plain}
% \bibliography{bibliography}

\end{document}